# SharkTrack: an accurate, generalisable software for streamlining shark and ray underwater video analysis

Filippo Varini[1,2], Joel H. Gayford[1,2,4], Jeremy Jenrette[3], Matthew J. Witt[6], Francesco Garzon[6;], Francesco Ferretti[3], Sophie Wilday[9], Mark E. Bond[5], Michael R. Heithaus[5], Danielle Robinson[10], Devon Carter[7], Najee Gumbs[8], Vincent Webster[8], Ben Glocker[1]

## Institutions

[1] Imperial College London, South Kensington Campus, London SW7 2AZ, United Kingdom

[2] Shark Measurements, London, United Kingdom

[3] Department of Fish and Wildlife Conservation, Virginia Tech, Cheatham Hall, 310 West Campus Drive, Blacksburg, VA 24061, USA

[4] College of Science and Engineering, James Cook University, 1 James Cook Drive, Douglas, Townsville QLD 4811, Australia

[5] Florida International University, 11200 SW 8th St, Miami, FL 33199, USA

[6] University of Exeter, Stocker Road, Exeter EX4 4PY, United Kingdom

[7] Anguilla National Trust, The Valley, Anguilla

[8] Fisheries, Department of Natural Resources, Government of Anguilla, Anguilla

[9] Bangor University, Bangor, Gwynedd LL57 2DG, United Kingdom

[10] School of Earth and Environment, University of Leeds, Leeds, LS2 9JT

## Correspondence

Filippo Varini,

Piazzetta Solofrano 9, 80123, Napoli, Italy

fppvrn@gmail.com

## Acknowledgements

We acknowledge the Paul G. Allen family foundation for funding the Global Finprint Project, which provided part of the data to train the model. FF acknowledges the financial support of the Bertarelli Foundation. We thank Dr Fabio De Sousa Ribeiro, Rajat Rasal, and Orlando Timmerman for providing guidance throughout the project. We thank students Natalie Ng, Rui Wen Lim, Michael Sellgren, Lara Tse, Steven Chen, Maria Pia Donrelas, Manfredi Minervini, and Xuen Bei (Bay) Chin for contributing to preliminary data annotation. We thank Adam Whiting, Aurora Crocini, Gabriele Bai, and Stephanie Guerinfor for supporting the evaluation of SharkTrack. BRUVS data from Anguilla were made possible through the UK Government's Biodiversity Challenge Funds Darwin Plus Initiative (DPLUS136) awarded to the Marine Conservation Society (UK), Anguilla National Trust, Department of Natural Resources – Anguilla and the University of Exeter (UK).

## Abstract

Elasmobranchs (shark sand rays) represent a critical component of marine ecosystems. Yet, they are experiencing global population declines and effective monitoring of populations is essential to their protection. Underwater stationary videos, such as those from Baited Remote Underwater Video Stations (BRUVS), are critical for understanding elasmobranch spatial ecology and abundance. However, processing these videos requires time-consuming manual analysis that can delay conservation. To address this challenge, we developed SharkTrack, a semi-automatic underwater video analysis software. SharkTrack uses Convolutional Neural Networks (CNN) and Multi-Object Tracking to automatically detect and track elasmobranchs and provides an annotation pipeline to manually classify elasmobranch species and compute species-specific

MaxN (ssMaxN), the standard metric of relative abundance. When tested on BRUVS footage from locations unseen by the CNN model during training, SharkTrack computed ssMaxN with 89% accuracy over 207 hours of footage. The semi-automatic SharkTrack pipeline required two minutes of manual classification per hour of video, an estimated 95% reduction of manual analysis time compared to traditional methods. Furthermore, we demonstrate SharkTrack's accuracy across diverse marine ecosystems and elasmobranch species, an advancement compared to previous models, which were limited to specific species or locations. SharkTrack's applications extend beyond BRUVS, facilitating the analysis of any underwater stationary video. By making video analysis faster and more accessible, SharkTrack enables research and conservation organisations to monitor elasmobranch populations more efficiently, thereby improving conservation efforts. To further support these goals, we provide public access to the SharkTrack software.

## 1. Introduction

Elasmobranch populations are experiencing global declines despite their ecological importance (Heithaus et al., 2022; Dulvy et al., 2021). Effective conservation strategies require accurate assessments of their spatial ecology and relative abundance (Kanive et al., 2021). Underwater stationary videos, such as those from Baited Remote Underwater Video Stations (BRUVS), are widely used non-invasive tools for monitoring elasmobranchs. BRUVS are underwater cameras equipped with bait to attract marine life, allowing researchers to monitor species abundance and behaviour (Harvey et al., 2013). However, the primary limitation of BRUVS and other underwater stationary video methods is the labour-intensive manual analysis required to process the footage. Researchers must analyse hours of video manually and this time-consuming process limits sample sizes and can introduce biases (Florisson et al., 2018), underscoring the need for automated solutions.

Machine Learning (ML) offers a promising solution to the video analysis bottleneck. Convolutional Neural Networks (CNNs) can be trained to locate and classify elasmobranch species, removing the need to watch hours of video footage. A promising effort is the Shark Detector which achieves 92% accuracy in determining whether an image contains a shark and can classify 67 shark species and 22 genera, representing 12.5% and 21.5% of all known species and genera, respectively (Jenrette et al., 2022, Jenrette et al., unpublished data). However, expanding this model to classify every elasmobranch species requires extensive collection of visual data from diverse environments, which might be unavailable or expensive to acquire. This limitation significantly hinders the broader application of elasmobranch detection models, constraining them to specific locations and species (Peña et al., 2020; Ulloa et al., 2020; Merencilla et al., 2021; Villon et al., 2024).

Here, we present a novel approach to achieving universal geographic and taxonomic elasmobranch classification by combining human expertise with machine learning. We introduce SharkTrack, a semi-automatic software for analysing underwater elasmobranch videos. SharkTrack takes raw footage as input and outputs the species-specific MaxN (ssMaxN), the maximum number of individuals of a species observed in a single frame. We test the speed, accuracy, and generalisability of SharkTrack on a diverse set of BRUVS videos, demonstrating that the software reduces video analysis time by 95% while retaining high ssMaxN accuracy and generalising across geographic locations and species. We provide SharkTrack as open-source software to enable researchers globally to streamline their data analysis, thus contributing to the conservation efforts of elasmobranch populations.

## 2. The SharkTrack Pipeline

SharkTrack is a single-class CNN that detects all elasmobranchs in a video without species-specific classification. This detection model is paired with an annotation pipeline that tracks each detected elasmobranch with Multi-Object Tracking (MOT) and allows users to manually classify its species. SharkTrack is designed to run on a standard laptop, without requiring specialised programming skills or a dedicated Graphics Processing Unit (GPU).

SharkTrack is available on GitHub (https://github.com/filippovarini/sharktrack) with instructions for users to analyse their underwater videos. Briefly, users can download the SharkTrack software and install the requirements saved in the ./requirements.txt file. Next, they can run the elasmobranch detector on their videos using the ./app.py file (Fig. 1a,b). After running the model, the output will include one image for each tracked elasmobranch (Fig. 1c,d). Users can visualise all detection images, delete invalid detections and assign the species to the appropriate ones by renaming the file (Fig. 1e). Finally, the MOT software transfers the annotations back to the raw output and computes the final MaxN (Fig. 1f).

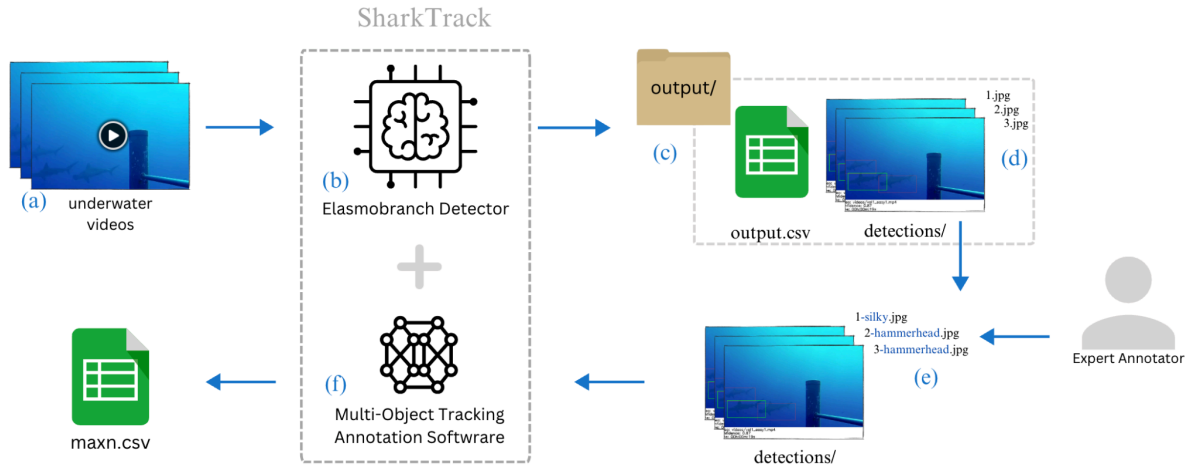

*Figure 1: The SharkTrack AI-enhanced underwater video analysis pipeline. The user collects a set of BRUVS video footage (a) and then runs the SharkTrack detector on the videos (b). The detector saves all detection information in a CSV file (c) and stores a detection image for each individual elasmobranch (d). The user manually annotates the species of each detection by renaming the file (e). SharkTrack updates all the sightings of each elasmobranch tracked with MOT based on the expert classification and computes MaxN (f).*

2.1 Data Collection

To train a model robust to a variety of geographic locations and elasmobranch species, we compiled a taxonomically and geographically diverse elasmobranch image dataset. We compiled 77 videos of elasmobranchs, totalling approximately 2 hours of footage. These videos are snippets of varying duration extracted by their owners from videos collected in 25 different locations around the world BRUVS (Fig. 2).

We extracted one frame per second from the videos and skilled personnel, capable of identifying elasmobranch species (JG, SL), annotated the images by drawing bounding boxes around all elasmobranchs seen in each frame and classifying their species. The resulting dataset contained

6.5 times more annotations of superorder Selachii (sharks) than Batoidea (rays). As this class imbalance could negatively affect the model performance (Schneider et al., 2020), we added 587 more instances of Batoidea, sourced from social media, which are openly available (Kuznetsova et al., 2020). We further preprocessed the images to remove biases and inaccuracies, as described in Supporting Information S1, producing a final dataset of 6,800 images with bounding box annotations.

We split the dataset into training and validation sets, using a random 85:15 split on videos rather than images, to ensure full independence between images of the training and validation set. Finally, to test the ability of SharkTrack to detect elasmobranchs in new locations, we collected 207 hours of BRUVS footage from three locations not represented in the training data (Table 1, Williams et al., 2024, Robinson 2022), which had already been manually analysed and annotated with the ssMaxN metric.

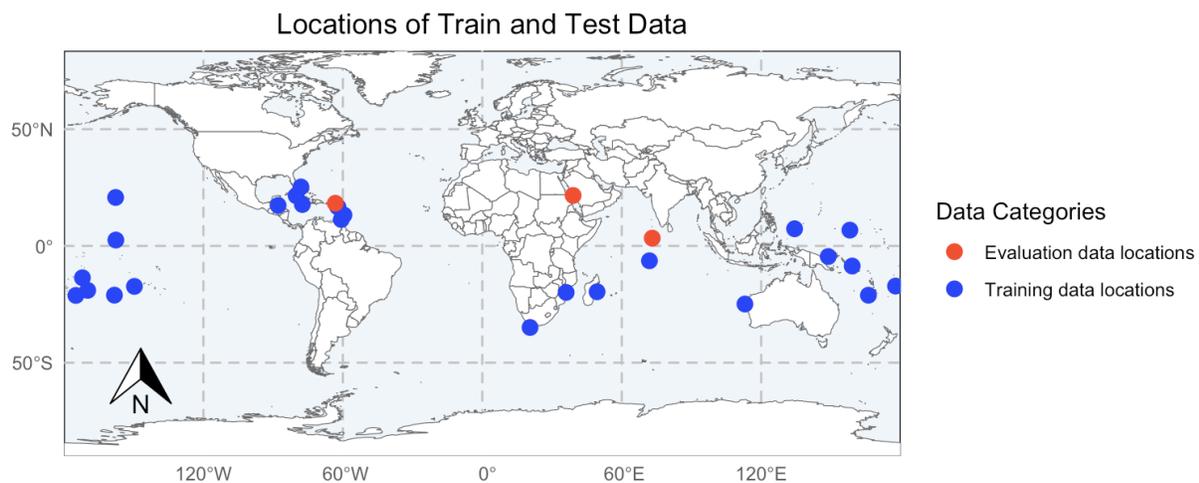

*Figure 2: The training dataset consists of 6,862 images extracted from 77 BRUVS sourced from 25 locations worldwide. The evaluation dataset compiles 207 hours of BRUVS videos collected in 3 out-of-sample locations.*

*Table 1: Metadata of out-of-sample BRUVS videos included in the evaluation dataset.*

| Location | Red Sea | Anguilla | Maldives |
|---|---|---|---|
| **Total footage from multiple BRUVS (hours)** | 98 | 63.5 | 45.5 |
| **Deployment type** | Benthic | Benthic | Benthic |
| **Species count** | 15 | 5 | 7 |

2.2 Elasmobranch Detector

The elasmobranch detector was built using the YOLO (You Only Look Once) architecture (Redmon et al., 2016), a deep-learning CNN model designed for real-time object detection. YOLO processes entire images in a single pass, predicting both object locations and classes simultaneously, making it faster than many traditional methods. We implemented the YOLOv8 version, provided by Ultralytics (Jocher et al., 2023), which offers models of varying sizes, from 'nano' to 'extra large.' These model sizes reflect a tradeoff between speed and accuracy, where smaller models (like 'nano') are faster but may be less accurate, while larger models offer higher accuracy but require more computational resources and time.

We evaluated different model versions using the mean average precision at an Intersection over Union (IoU) threshold of 50% (mAP50). The mAP50 is a performance metric for object detection models that calculates the average precision of prediction bounding boxes, which overlap at least 50% of the ground truth (additional details on used performance metrics are in Supporting Information S2).

To create a robust binary elasmobranch classifier, we clustered the species-level annotations into a single 'elasmobranch' class and applied data augmentation strategies suggested for object detectors (Rotation, Equalisation, Horizontal Flipping, Zoph et al., 2020). We trained two small model sizes, the 'nano' and 'medium' versions, for 500 epochs with a patience of 50, a batch size of 64, a confidence threshold of 0.2 and an image size of 640. The nano and medium models achieved a mAP50 score of 0.85 and 0.86 and an F1 score of 0.83 and 0.84, respectively. Since the nano model ran three times faster, we chose this as the final model.

2.3 Multi-Object Tracker (MOT)

We developed an annotation software component that allows experts to annotate the detected elasmobranch species, leveraging the MOT software BotSORT (Aharon et al., 2022). This component tracks each elasmobranch with a consistent track ID as long as it remains within a continuous video segment. If the same individual leaves the segment, the tracker forgets it and assigns a new track ID if it re-enters the video. MOT enables researchers to classify each track ID only once, updates the classification of all detections with the same track ID, and deletes all sightings of a false positive detection, such as a fish or algae. The annotation software updates the model output with these manual annotations to compute the ssMaxN.

We tuned the parameters of the BotSORT tracker and developed a postprocessing algorithm to improve the tracking accuracy in underwater environments, as described in Supporting Information S3.

## 3. Evaluation on New Locations

We evaluated SharkTrack on three case studies of elasmobranch videos collected from novel locations not represented in the training data: the Red Sea, the Caribbean Sea, and North Malé Atoll in the Maldives (Fig. 2). We compared the ground truth ssMaxN, computed manually, with the ssMaxN computed by analysts using the SharkTrack semi-automatic process and evaluated the SharkTrack MaxN Accuracy metric, proposed by Villon et al. (2024). For each video $i$, the MaxN Accuracy is defined as:

$$MaxN\ Accuracy_i = \frac{Correct\ MaxN\ Prediction_i}{Correct\ MaxN\ Prediction_i + Incorrect\ MaxN\ Prediction_i}$$

Where $Correct\ MaxN\ Prediction_i$ and $Incorrect\ MaxN\ Prediction_i$ are the number of species for which the predicted ssMaxN in the analysed video was correct and incorrect, respectively. We also evaluated the processing time required by the ML model (ML Inference Time) and the additional time needed to manually post-process the SharkTrack output to compute ssMaxN (Manual Post-Processing Time, Fig. 1c,d,e). To evaluate the reduction in manual analysis, we compared the Manual Output Post-Processing Time to the time required by experts to manually analyse the footage using traditional methods without SharkTrack (Traditional Manual Analysis Time). Additional details about the process we followed to evaluate the time are provided in Supporting Information S4.

Across the three case studies, SharkTrack achieved an average of 89% MaxN Accuracy (Table 2). This metric was computed over 153 MaxN values found in 207 hours of footage. SharkTrack also found the MaxN manually computed by experts to be incorrect in 7% of the instances, due to observers not detecting elasmobranchs present in the videos (e.g., expert MaxN = 0, correct

MaxN = 1, or expert MaxN < correct MaxN; Fig. 4c; Fig. 5b). By using SharkTrack, the same analysts could detect the previously missed elasmobranch, thus reducing human error.

*Table 2: Summary results of the case studies on which SharkTrack was tested.*

| Location | Red Sea | Anguilla | Maldives |
|---|---|---|---|
| **Total footage from multiple BRUVS (hours)** | 98 | 63.5 | 45.5 |
| **SharkTrack ML Inference Time (hours)** | 32.6 | 48.7 | 9.1 |
| **Manual Output Post-Processing time (hours)** | 2.9 | 1.5 | 2.4 |
| **Traditional Manual Analysis Time (without SharkTrack, hours)** | 65.3 | 47.6 | 30.3 |
| **Number of MaxN Indices** | 50 | 32 | 71 |
| **Number of MaxN Indices Corrected** | 0 | 2 | 8 |
| **Average MaxN Value** | 1 | 1 | 2 |
| **MaxN Accuracy** | *81.8%* | *87.8%* | *97.8%* |

As shown in Table 3, across the three case studies, on a 6-Core Intel Core i7 CPU with 16GB of RAM, SharkTrack processed the videos automatically in 27 minutes per video hour, on average.

The model output then required an average of 2 minutes of manual post-processing per video hour, to compute the final ssMaxN. Compared with the traditional manual analysis time without SharkTrack, which was 42 minutes per video hour on average, SharkTrack reduced the time required for manual analysis by 95%. The 27-minute inference time for SharkTrack is not included in this comparison, as it runs automatically in the background without requiring user intervention, unlike manual analysis.

Table 3: Time evaluation of each of the three case studies after being divided by the total footage for the corresponding case study, for normalisation.

| **Location** | *Red Sea* | *Anguilla* | *Maldives* | **Average** |
| --- | --- | --- | --- | --- |
| **Total footage from multiple BRUVS (hours)** | 98 | 63.5 | 45.5 | 69 |
| **SharkTrack ML Inference Time (minute per video-hour)** | 20 | 46 | 16 | 27 |
| **Manual Output Post-Processing time (minute per video-hour)** | 1.8 | 1.5 | 3.2 | 2 |
| **Traditional Manual Analysis Time (minute per video-hour)** | 40 | 45 | 40 | 42 |

Below, we analyse how the different environmental conditions of each case study affected the software's accuracy and speed.

## 3.1 Red Sea

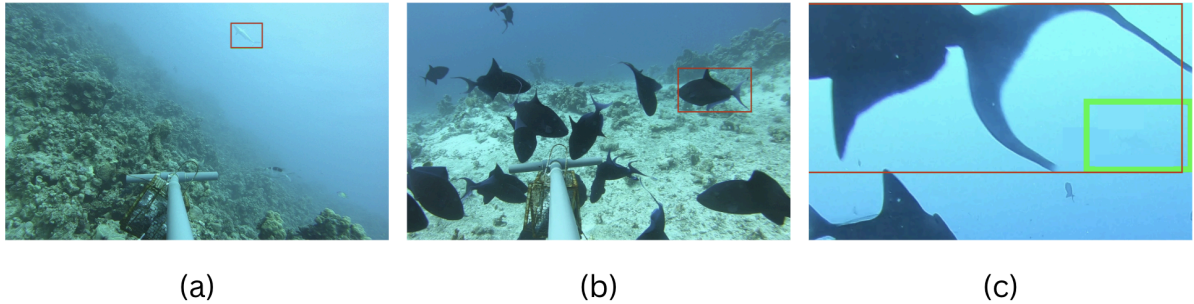

(a) (b) (c)

*Figure 3: Sample detections from the Red Sea dataset. The red bounding boxes are predicted by the model and the green is manually drawn by experts. The model correctly detected distant elasmobranchs (a) but occasionally confused reef fish (b) and missed elasmobranchs hidden by reef fish (c, in green).*

The videos collected in the Red Sea presented rocky reef habitats with clear water. The visibility was more than 5 metres in most videos and allowed the SharkTrack model to detect elasmobranchs at a greater distance than the other regions (Fig. 3a), produce a lower number of false positive detections, and achieve a lower ML inference time than average (Table 2). However, the video would occasionally be cluttered by schools of reef fish, causing false positive detections (Fig. 3b). In these cases, the model failed to detect distant elasmobranchs obscured by reef fish (Fig. 3c), lowering the MaxN Accuracy.

## 3.2 Caribbean

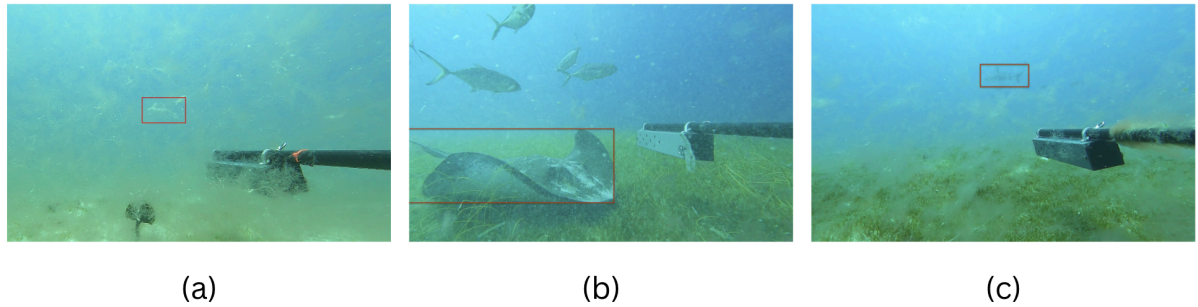

(a)　　　　　　　　　　　　(b)　　　　　　　　　　　　(c)

*Figure 4: Sample detections from the Caribbean evaluation dataset. All the bounding boxes are predicted by the model. The videos presented turbid water obscured by algae, which the model often confused (a), while still correctly detecting most elasmobranchs (b)(c).*

The BRUVS footage collected in the Caribbean Sea presented a seagrass habitat with visibility often obscured by floating algae. In this turbid water, SharkTrack confused many of the algae as elasmobranchs (Fig. 4a). About 95% of model detections were false positives, which caused a longer ML inference time. In contrast, algae were easy to differentiate from elasmobranchs in the Output Post-Processing stage, thus analysts could delete false detections faster than the other regions (Table 2). The low number of elasmobranchs to classify, maximum one individual per frame, also contributed to faster analysis time. Despite the turbid water, SharkTrack achieved an accuracy of 87.8%. SharkTrack also detected elasmobranchs even when they were poorly visible or occluded by floating algae (Fig 4b,c) and were not detected by human analysts alone (Fig. 4c).

## 3.3 Maldives

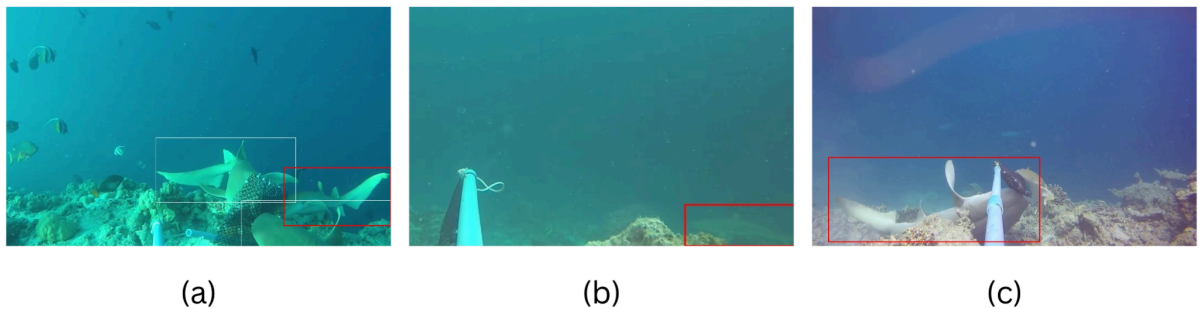

(a) (b) (c)

*Figure 5: Sample detections from the Maldives evaluation dataset. All the bounding boxes are predicted by the model. The videos presented high elasmobranch abundance (a), occasionally not spotted by human observers (b) and assuming twisted positions in the frame (c).*

The BRUVS footage collected in the Maldives comprised rocky reefs with high water quality and visibility. As few reef fish obscured the video, SharkTrack achieved the lowest rate of false positive detections and, thus, the fastest ML inference time (Table 2). The videos also showed the highest shark abundance, with up to three individuals per frame (Fig. 5a). Due to more detections requiring manual species classification, the Maldivian videos required the highest manual analysis time, more than double the time needed for the Caribbean case (Table 2). SharkTrack detected elasmobranchs in the distance that were not found by human observers alone (Fig. 5b) and achieved the highest ssMaxN accuracy of 97.8%. SharkTrack also incorrectly predicted two ssMaxN indices because it failed to detect sharks that were twisting their bodies (Fig. 5c).

It is important to note that SharkTrack was trained on video collected from the same country, the Maldives, although from different sites to the North Malé Atoll, where the 45.5 hours of footage used to evaluate SharkTrack were obtained. Although the training sites are more than 50 km from the North Malé Atoll, they may share similar environmental backgrounds and species compositions. We believe this similarity contributes to the ML detector´s high accuracy in this case study, highlighting the advantage of training underwater ML detectors with videos that closely resemble those they will eventually analyse.

## 4. Discussion

Here, we presented SharkTrack, an AI-enhanced semi-automatic software that reduces by an average of 95% (standard deviation, sd: 0.02) the manual analysis time of underwater video of elasmobranchs while producing abundance indices with an average of 89% (sd: 8.1) accuracy, across locations, and species. This analytical pipeline leverages researchers' expertise to classify elasmobranchs while reducing analysis time and removing human error caused by missed detections (Table 2). Furthermore, SharkTrack runs seamlessly on a basic laptop commonly used in the field, a portability that makes it accessible for researchers in resource-limited settings.

SharkTrack was developed using different methods than previous efforts, which generally trained ML models focused on a narrow set of species and locations. For example, Villon et al., (2024) developed a Faster R-CNN shark detector to compute MaxN from BRUVS in New Caledonia but limited detection and evaluation to three of the seven shark species in the area, achieving an average F1 score of 0.67. This model does not detect all elasmobranch species found in BRUVS and does not replace manual analysis for extracting broader community indices. Here, we wanted to increase the taxonomic scope, speed, and performance of these approaches to develop a universal species classifier. While ongoing efforts work on collecting and aggregating existing data from various sources to achieve this goal (Jenrette et al. 2022, Jenrette et al., unpublished data), our approach achieves geographical and taxonomic generalisation without requiring further data collection, by leveraging human-machine interaction.

SharkTrack can produce many false positive detections (up to 95%), mainly in new benthic ecosystems. This limitation affects ML models when applied to new, highly variable natural habitats (Kellenberger et al. 2019). These errors must be manually removed in the manual post-processing stage. With the MOT annotation platform, users can remove all sightings of a tracked false positive detection by deleting the related image from the output directory. This ensures minimal delay and maintains a fast analysis time, as demonstrated in Table 2. Yet, the manual species classification stage is still a significant bottleneck. Analysing videos with an average MaxN of two takes twice as long as those, with a MaxN of one. We did not have videos with greater MaxNs, but expect longer times for videos with higher elasmobranch abundances, highlighting the increased manual effort required as the number of detected individuals grows (though this is not a unique challenge for SharkTrack; high abundance leads to higher manual processing time even for humans). Further research is needed to better quantify the time savings provided by SharkTrack, in relation to the elasmobranch abundance.

Although in this study we evaluated SharkTrack on BRUVS footage, the software can process any elasmobranch underwater stationary video. BRUVS are commonly used because they reduce the amount of video time needed to detect present individuals since they are attracted by the bait. However, BRUVS influences elasmobranch behaviour and, at high densities, shows a nonlinear relationship between MaxN (maximum number observed) and true density, making it difficult to detect abundance variations (Harvey et al., 2018). A potential solution to these issues is to use longer, unbaited video recordings, which would offer a more accurate representation of elasmobranch behaviour and population density. Because SharkTrack significantly streamlines the video analysis process, it makes it feasible for researchers to adopt these longer, unbaited recordings, allowing for better population estimates without greatly increasing the time required for analysis.

SharkTrack also contributes to the ML community by accelerating future developments of species-specific classifiers. The model extracts elasmobranchs automatically from unlabelled footage, and the MOT annotation software allows developers to rapidly generate a dataset of elasmobranch images with bounding box species classifications, which can be used to train object detection models. In the future, SharkTrack could be integrated with species-specific classifiers, such as the Shark Detector, and automatic size estimation algorithms from stereo video (Boër et al., 2023). This would enable the development of a fully automatic software capable of analysing underwater stationary videos, providing both abundance and size estimates for elasmobranchs and aiding their conservation further.

# Supplementary Material for SharkTrack: an accurate, generalisable software for streamlining shark and ray underwater video analysis. Varini et al. 2024, Methods in Ecology and Evolution.

## S1 Image Inpainting of Training Data

Some videos had logo and white text annotations of the species visible within the image. Because CNNs can focus on text rather than images (Lapuschkin et al 2019), we covered these annotations by adding black patches to the images, a method known as image inpainting. This technique involves two main steps: detecting pixels belonging to text and logos and subsequently covering them with black patches. Given that all annotations were bright white, we identified text and logos by converting their RGB values to grayscale and isolating pixels with brightness values above 230. We then calculated the coordinates of the rectangles enclosing each letter and overlaid a black box on these areas.

Occasionally, other parts of the image, such as shark bellies, were as bright and were incorrectly identified as text. However, this did not negatively impact model training, as the black boxes can be considered a form of Cutout image augmentation (DeVries et al. 2017), which is known to enhance model robustness. To minimise the occurrence of false text detection, we selected a brightness threshold of 230.

## S2 Object Detection Evaluation Metrics

We evaluated different model versions using the mean average precision at an Intersection over Union (IoU) threshold of 50% (mAP50). The mAP50 is a performance metric for object detection models that calculates the average precision of prediction bounding boxes which overlap at least 50% of the ground truth. Some previous CNN detectors mentioned are evaluated with the F1 score, which provides a balanced view of the model's overall accuracy by considering both false positives (incorrectly identified objects) and false negatives (missed detections). These metrics are expressed mathematically as:

$$mAP50 = \frac{1}{N} \sum_{i=1}^{N} AP_i,$$

$$F1 = 2 \cdot \frac{Precision \cdot Recall}{Precision + Recall},$$

$$Precision = \frac{TP}{TP+FP}, \text{ and}$$

$$Recall = \frac{TP}{TP + FN}.$$

where TP, FP, FN represent the number of true positives, false positives and false negatives, respectively.

## S3 Multi-Object Tracker Fine-Tuning and Post-Processing Algorithm

The MOT software employed a BotSORT tracker. BotSORT is a state-of-the-art MOT algorithm robust to camera motion and crowd occlusion. To fine-tune the tracker to the underwater environment, we performed a grid search on its hyperparameters. We evaluated each configuration with the TrackEval software (Luiten & Hoffhues 2020) and optimised the MOTA evaluation metric (Bernardin et al 2006, May).

MOTA, or Multiple Object Tracking Accuracy, is a comprehensive metric used to evaluate the performance of tracking algorithms. It combines three types of errors: false positives (incorrectly detected objects), false negatives (missed objects), and identity switches (instances where the track ID is incorrectly assigned). A higher MOTA score indicates better tracking performance by minimising these errors.

The best configuration achieved a MOTA of 77%, comparable to the state-of-the-art MOT in less three-dimensional domains like pedestrian tracking. A tracker needs detections at a higher frame rate per second (FPS) rate to be more accurate. However, a higher FPS causes slower inference time. We calculated the tradeoff between MOTA accuracy and inference speed on the validation data and chose an FPS rate of 3 frames per second as it provided the best tradeoff.

Although the model robustly detects elasmobranchs in new locations, the highly varying marine habitats cause it to output many false positive detections. To solve this issue, we developed a postprocessing algorithm that predicts if a track is false positive and deletes all the associated detections. The postprocessing algorithm selects all detected tracks that last for less than one second or whose centre moves by less than 0.08%. The algorithm then removes all selected tracks whose maximum detection confidence is lower than 0.7. We selected these values through a grid search process. Using these optimal hyperparameters, the postprocessing algorithm removes 40% of false positive detections while sacrificing less than 0.08% of true positives and reduces the number of detections the user needs to delete with this MOT annotation platform manually.

# S4 Time Evaluation of SharkTrack

We evaluated the ML Inference Time by running the SharkTrack model for each use case on the same machine, a 2019 Macbook Pro with a 2.6 GHz 6-Core Intel Core i7 CPU and 16GB of RAM. For each use case, we measured inference time with the "time" Unix command, as given by the "real" value.

To measure Manual Post-Processing Time, we first uploaded the model detections to the cloud for each use case. Next, a biologist analysed all detections as required by SharkTrack to compute MaxN and recorded the time taken for each use case.

In each of the three case studies used to evaluate SharkTrack, an elasmobranch expert manually analysed the footage using the traditional method to compute ssMaxN. The experts also estimated the total time it took them to complete this manual analysis, and we recorded this time as the Traditional Manual Analysis Time.

We measured the